\documentclass[sigplan,nonacm]{acmart}
\setcopyright{acmlicensed}
\copyrightyear{2025}
\acmYear{2025}
\acmDOI{XXXXXXX.XXXXXXX}

\usepackage{pifont}
\usepackage{listings}
\usepackage{subcaption}
\usepackage{multirow}
\usepackage{soul}
\usepackage{xspace}
\usepackage{enumitem}
\usepackage[ruled,vlined,linesnumbered]{algorithm2e}

\def\BibTeX{{\rm B\kern-.05em{\sc i\kern-.025em b}\kern-.08em
    T\kern-.1667em\lower.7ex\hbox{E}\kern-.125emX}}

\newlist{todolist}{itemize}{2}
\setlist[todolist]{label=$\square$}
\usepackage{pifont}
\newcommand{\GrayCircle}[1]{\raisebox{-0.3ex}{\text{\LARGE\color{darkgray}#1}}}
\newcommand{\cOne}{\GrayCircle{\ding{182}}} 
\newcommand{\cTwo}{\GrayCircle{\ding{183}}} 
\newcommand{\hive}{\emph{Hive}}
\newcommand{\LogitsCache}{\textsc{Logits Cache}}
\newcommand{\AAS}{\textsc{Agent-Aware Scheduling}}
\graphicspath{{./images/}} 

\begin{document}

\title{Hive: A Multi-Agent Infrastructure for Algorithm- and Task-Level Scaling}

\author{Zizhang Luo}
\authornote{Both authors contributed equally to this work.}
\email{semiwaker@pku.edu.cn}
\orcid{0000-0002-7276-2317}
\affiliation{%
  \institution{Peking University}
  \city{Beijing}
  \country{China}
}

\author{Yuhao Luo}
\authornotemark[1]
\email{luoyuhao584@gmail.com}
\affiliation{%
  \institution{Peking University}
  \city{Beijing}
  \country{China}
}

\author{Youwei Xiao}
\email{shallwe@pku.edu.cn}
\affiliation{%
  \institution{Peking University}
  \city{Beijing}
  \country{China}
}

\author{Yansong Xu}
\email{yansongxu.mail@gmail.com}
\affiliation{%
  \institution{Peking University}
  \city{Beijing}
  \country{China}
}

\author{Runlin Guo}
\email{linyu22373@gmail.com}
\affiliation{%
  \institution{Peking University}
  \city{Beijing}
  \country{China}
}

\author{Yun Liang}
\email{ericlyun@pku.edu.cn}
\orcid{0000-0002-9076-7998}
\affiliation{%
  \institution{Peking University}
  \city{Beijing}
  \country{China}
}

\begin{abstract}

Large language models are increasingly deployed as complex agentic systems that scale with task complexity. While prior work has extensively explored model- and system-level scaling, algorithm- and task-level scaling remain largely unaddressed, constraining the full potential of agentic systems. At the algorithm level, allocating additional inference-time computation can enhance workflow capacity but introduces cross-path redundancy: overlapping computations across multiple reasoning branches. At the task level, complex tasks can be decomposed into subproblems and delegated across multiple agents for improved scalability and parallelism. However, existing infrastructures' scheduling is unaware of the existence of multiple agents, missing opportunities to optimize resource allocation.

We propose Hive, a multi-agent infrastructure that enables algorithm- and task-level scaling. Hive features a description frontend that captures per-agent behavior and supports test-time scaling algorithms. Leveraging this specification, our backend introduces two key mechanisms: \textbf{Logits Cache} that reuses intermediate logits across redundant sampling paths to mitigate cross-path redundancy at the algorithm level, and \textbf{Agent-Aware Scheduling} that efficiently allocates compute and KV-cache resources according to agent contributions at the task level. Experiments show that \textbf{Logits Cache} achieves an average speedup of $1.11\times$--$1.76\times$ for re-sampling, and \textbf{Agent-Aware Scheduling} reduces the hotspot miss rate by $33\%$--$51\%$.

\end{abstract}


\maketitle

\section{Introduction}

\begin{figure}[ht]
    \centering

    \begin{subfigure}[b]{\columnwidth}
    \centering
    \includegraphics[width=\columnwidth]{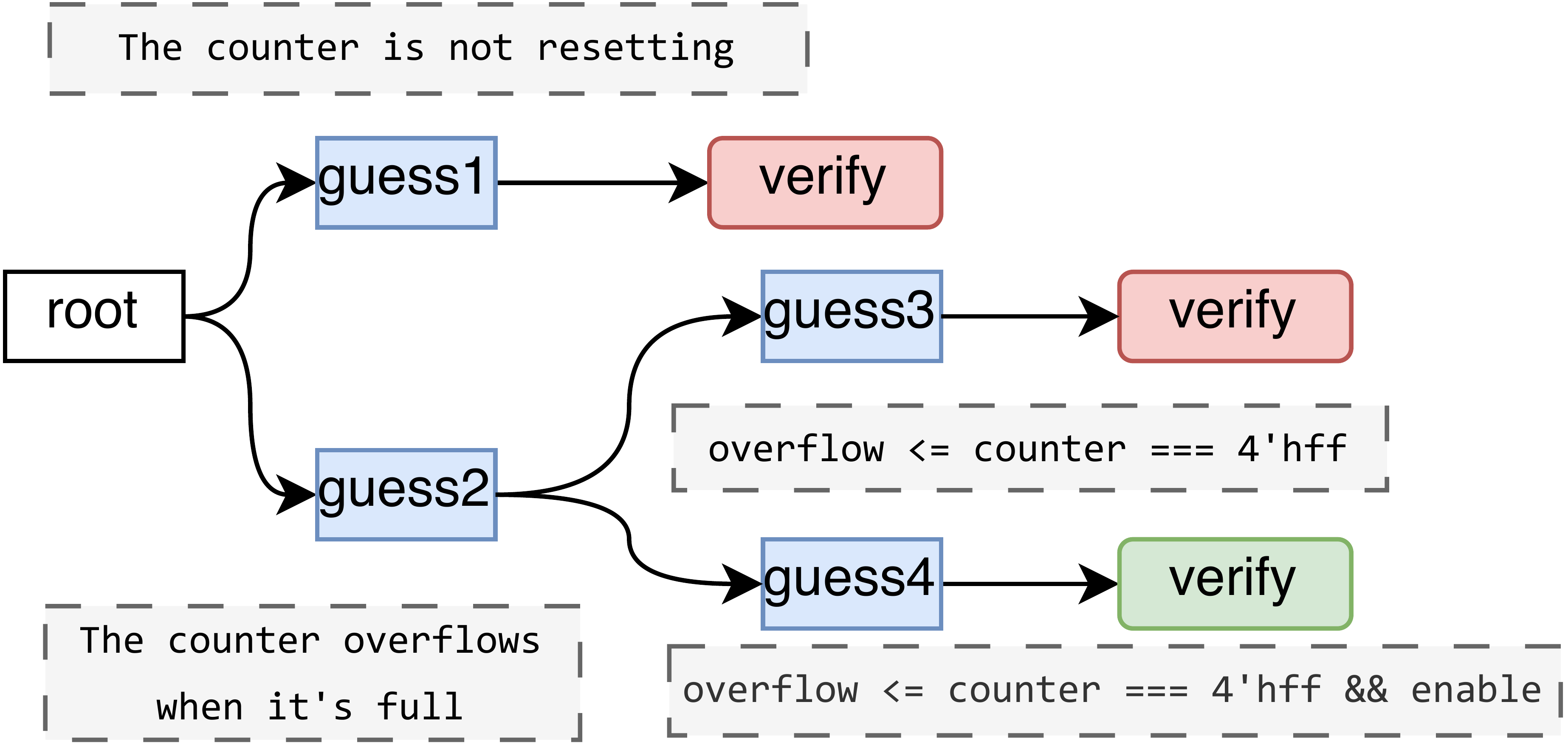}
    \caption{Algorithm-level Scaling using Tree-of-Thoughts}
    \label{fig:hw_verif_tot}
    \end{subfigure}
    

    \begin{subfigure}[b]{\columnwidth}
    \centering
    \includegraphics[width=0.9\columnwidth]{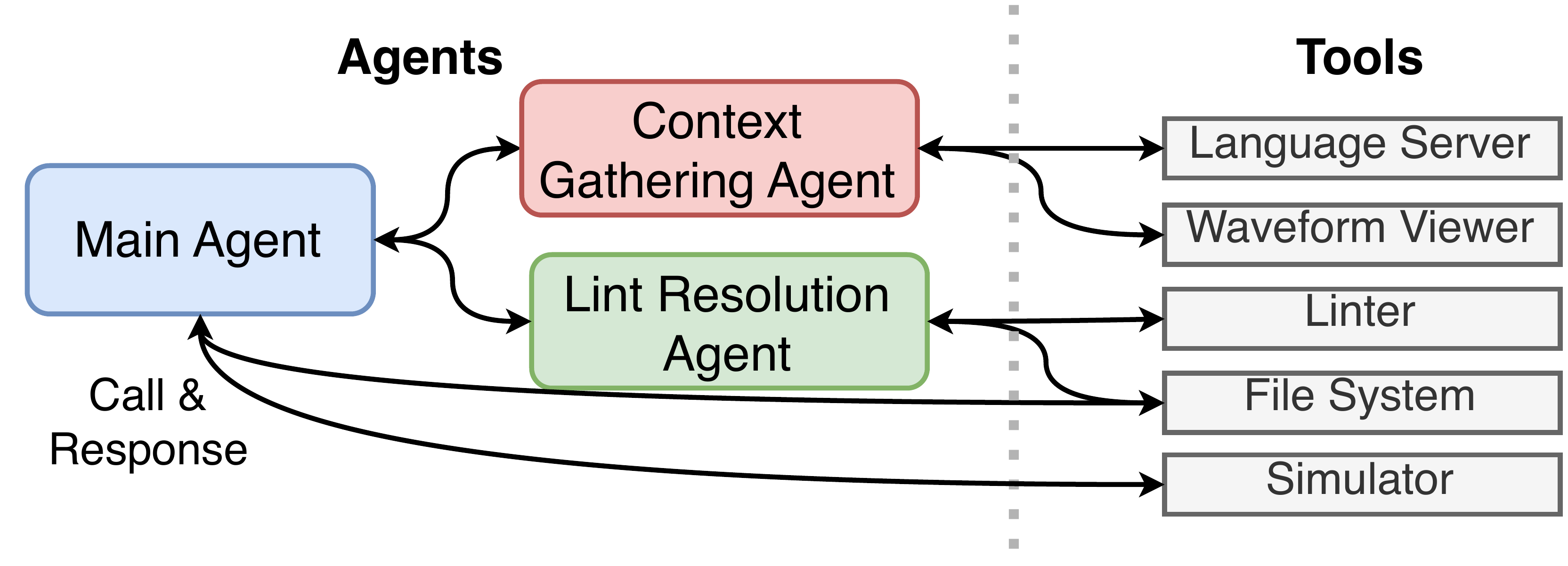}
    \caption{Task-level Scaling by Task Delegation}
    \label{fig:hw_verif_agent}
    \end{subfigure}

    \caption{Case study on a Hardware Verification Agent}
\end{figure}

Large language models (LLMs) have advanced rapidly in recent years \cite{kimi-K2.5, qwen3_coder_next, deepseekv32, openai_gpt54}, demonstrating strong capabilities in reasoning, generation, and decision-making. As their competence continues to grow, their role is also shifting from single-turn response engines to agentic problem solvers that can plan over multiple steps, invoke external tools, and coordinate toward complex goals\cite{openclaw_github, opencode, kimi_code}. In these emerging workloads, inference is no longer a collection of independent prompt-response pairs; instead, it becomes a stateful process in which context accumulates across turns, intermediate results are repeatedly reused, and multiple reasoning branches or agents may coexist within a single task. 


As problem complexity continues to grow, effectively solving real-world tasks often requires scaling LLM systems along multiple dimensions rather than relying on a fixed model and static inference procedure. To better characterize this trend, we summarize the scaling landscape of LLM systems into four complementary paradigms:

\begin{itemize}[leftmargin=*]
    \item \textbf{Model-Level Scaling} directly improves the capability of LLMs by increasing model size and fine-tuning with domain-specific datasets, typically leading to stronger reasoning and generation ability.
    \item \textbf{System-Level Scaling} improves serving efficiency through additional hardware resources and parallelism, enabling higher throughput, lower latency, and better scalability under concurrent workloads.
    \item \textbf{Algorithm-Level Scaling} improves the potential of the same workflow by allocating additional computation at inference time, e.g., test-time scaling\cite{tts-survey}.
    \item \textbf{Task-Level Scaling} improves the scalability and robustness of problem solving by decomposing complex objectives into multiple interacting subproblems, often coordinated through multi-agent systems.
\end{itemize}

However, optimization in existing inference systems has been concentrated mainly on model-level and system-level scaling. At the model level, substantial effort has been devoted to quantization and compression\cite{sageattention3, arkvale, qerl, efficientqat}, improving model-hardware efficiency by reducing memory footprint and computation cost. At the system level, a wide range of serving architectures and deployment strategies, such as PD-Disaggregation\cite{distserve, splitwise} and context parallelism\cite{ring-attn, ulysses}, have been developed to improve the utilization of available hardware resources. By contrast, little attention has been paid to runtime support for algorithm-level and task-level scaling, even though both are becoming increasingly important in general or domain-specific agentic workloads.


We illustrate the significance of algorithm- and task-level scaling through a case study on a hardware verification agent designed for RTL-level Automatic Program Repair (APR). 
At the algorithm level, the inherent stochasticity of large language models is unacceptable in high-assurance domains such as hardware verification, making test-time scaling essential for reliability. As shown in \autoref{fig:hw_verif_tot}, the agent employs Tree-of-Thought \cite{tree-of-thought} to navigate the solution space and identify a verifying patch. The low-level intricacies of hardware create a vast and sparse solution landscape, where a naive ReAct loop is prone to losing focus. In contrast, Tree-of-Thoughts enables efficient exploration across multiple reasoning branches. Yet, this approach introduces \textbf{Cross-Path Redundancy}: as branches expand, they often share common prefixes and intermediate generations, which is a universal challenge to the major test-time scaling methods based on re-sampling. Existing inference engines \cite{sglang, vllm} treat these branches as independent requests, leading to repeated computation and significant overhead.

Task-level scaling further enhances agentic systems by delegating subtasks to specialized agents. In the hardware verification example (\autoref{fig:hw_verif_agent}), three agents collaborate: a main agent orchestrates program repair and simulation, a context-gathering agent extracts information via language server and waveform viewer, and a lint-resolution agent handles static issues to offload the main agent. This orchestration keeps each agent's context focused and prevents overcrowding. However, multi-agent systems raise a new challenge for the infrastructure: \textbf{heterogeneous runtime characteristics} among agents—varying input/output lengths, cache footprints, and context reuse patterns. In this example, the main agent is more significant and takes many more tokens than the other two agents. Current inference engines remain largely agent-unaware, processing all requests uniformly and failing to exploit the distinct execution patterns of different agents.

In this paper, we present~\hive, a multi-agent inference infrastructure for hardware verification that rethinks how modern inference engines support algorithm-level and task-level scaling. The key insight behind~\hive~is that the inefficiency of scaling workloads does not merely arise from increased compute demand, but also from the fact that existing inference engines are not designed to fully exploit their structural regularities. \hive~is composed of a multi-agent description front-end and LLM inference back-end. The front-end captures the per-agent behavior, supports test-time-scaling algorithms, and models multi-agent interaction as asynchronous coroutine-like task spawning and synchronization.
In the back-end, for algorithm-level scaling,~\hive~introduces \LogitsCache, a new caching paradigm that reuses intermediate logits across redundant sampling paths. Unlike existing inference engines that treat different reasoning branches as independent executions, \LogitsCache~captures reusable probability distribution on states across branches and avoids repeated re-sampling computation, thereby accelerating hybrid test-time scaling while preserving output determinism. For task-level scaling, \hive~introduces the \AAS~mechanism that allocates compute and KV cache resources according to agent-specific runtime characteristics. By adapting resource allocation to the heterogeneous behavior of different agents, \hive~improves KV cache retention efficiency and overall serving utilization under multi-agent workloads. 

Experiments show that \LogitsCache~achieves an average speedup of $1.11\times$--$1.76\times$ for re-sampling, and \AAS~reduces the hotspot miss rate by $33\%$--$51\%$.



The main contributions of this paper are as follows:
\begin{itemize}[leftmargin=*]
    \item We propose~\hive, a multi-agent inference infrastructure for hardware verification that provides algorithm-level and task-level scaling by modeling multi-agent and test-time scaling with the front-end and jointly optimizes inference execution in the back-end.
    
    \item For \textbf{algorithm-level scaling}, we propose a new caching paradigm called \textbf{Logits Cache} for hybrid test-time scaling to reduce redundant computation.

    \item For \textbf{task-level scaling}, we identify distinctive heterogeneity in multi-agent workloads, and propose an agent-aware scheduler that allocates resources according to agent-specific characteristics.

\end{itemize}

\section{Background}

\begin{figure*}[t]
  \centering
  \includegraphics[width=\textwidth]{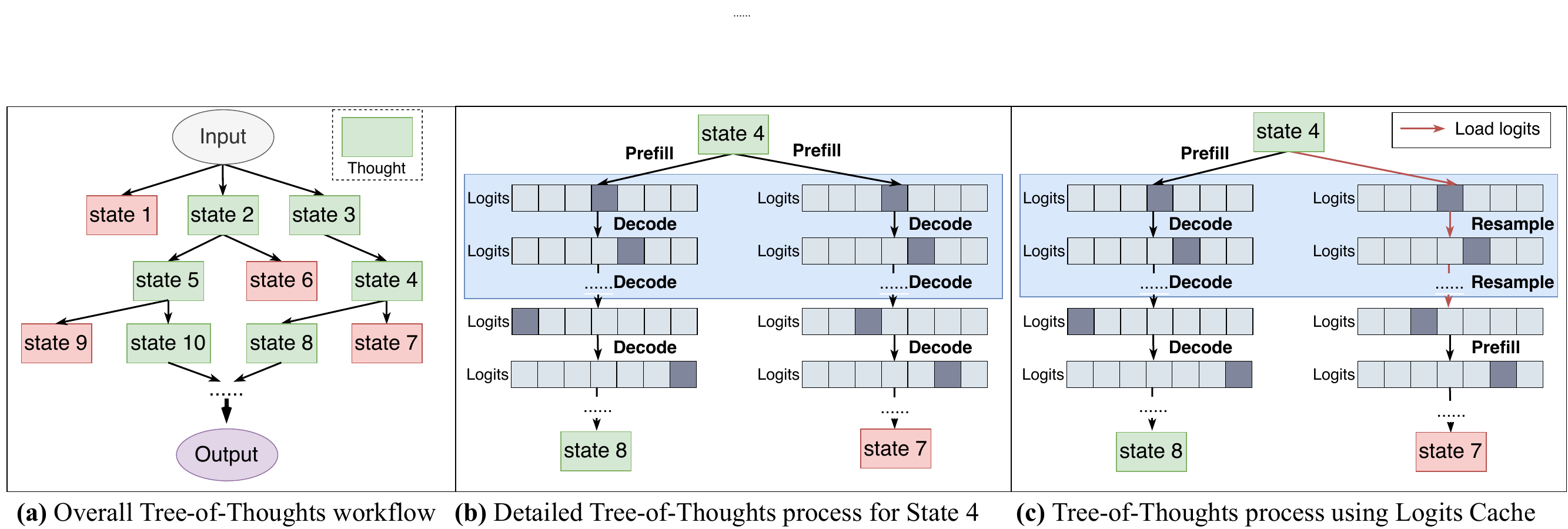}
  \caption{(a) An example of branching reasoning in Tree-of-Thought. (b) A detailed view of the sampling process for State 4. (c) Tree-of-Thoughts process using Logits Cache }
  \label{fig:tot_example}
\end{figure*}

\subsection{Large Language Model (LLM) Inference}
Modern LLMs\cite{kimi-K2.5, qwen3_coder_next, deepseekv32} are typically served through an autoregressive inference process. Given an input prompt, the serving system first executes a \emph{prefill} stage to process all prompt tokens and construct the KV cache, and then enters the \emph{decode} stage to generate output tokens one by one. At each decoding step, the model produces a logits distribution over the vocabulary, from which the next token is selected or sampled according to the decoding strategy. During decoding, the system reuses the cached keys and values from previous steps, thereby avoiding recomputation over the full prefix and significantly improving serving efficiency. This split execution model makes KV cache management, scheduling, and resource allocation central to the design of high-performance LLM inference systems\cite{distserve, sarathi-serve, splitwise}.


\subsection{Test-Time Scaling}
While training-time scaling has been the primary driver of recent advances in LLM capability\cite{traingModels, scaling_law}, its benefits have gradually slowed due to the high cost of pretraining and the limited availability of high-quality data. This trend has motivated the development of \emph{test-time scaling}\cite{tts-survey}, which aims to improve model performance by allocating additional computation during inference rather than during training. As an inference infrastructure, we focus on three types of test-time scaling: \emph{parallel}\cite{start, moa} scaling multi-samples the same prompt, \emph{sequential}\cite{self-refine, forest-of-thought} scaling iteratively improves on top of previous reply, and \emph{hybrid}\cite{react, tree-of-thought} scaling has both features of the prior two methods and forms a tree.


\subsection{Multi-Agent System}
A multi-agent system (MAS) consists of multiple interacting agents that collaborate through communication, information sharing, and decentralized decision-making to accomplish a common objective. MAS has been widely proposed as a pivotal pathway for harnessing collective intelligence while preserving the specialized characteristics of individual agents\cite{SurveyMAS, MultirobotSystems, exploringlargelanguagemodel}. Compared with a single-agent system, which handles the entire task within a single execution entity, MAS provides a more natural abstraction for complex tasks that require decomposition, role specialization, and iterative coordination, and more fine-grained context management. Besides, MAS also benefits from the parallel nature of LLMs, where batching can provide free computation and weight reuse, especially in self-hosted situations.



\section{Motivation}

\begin{figure}[t]
  \centering
  \includegraphics[width=0.9\columnwidth]{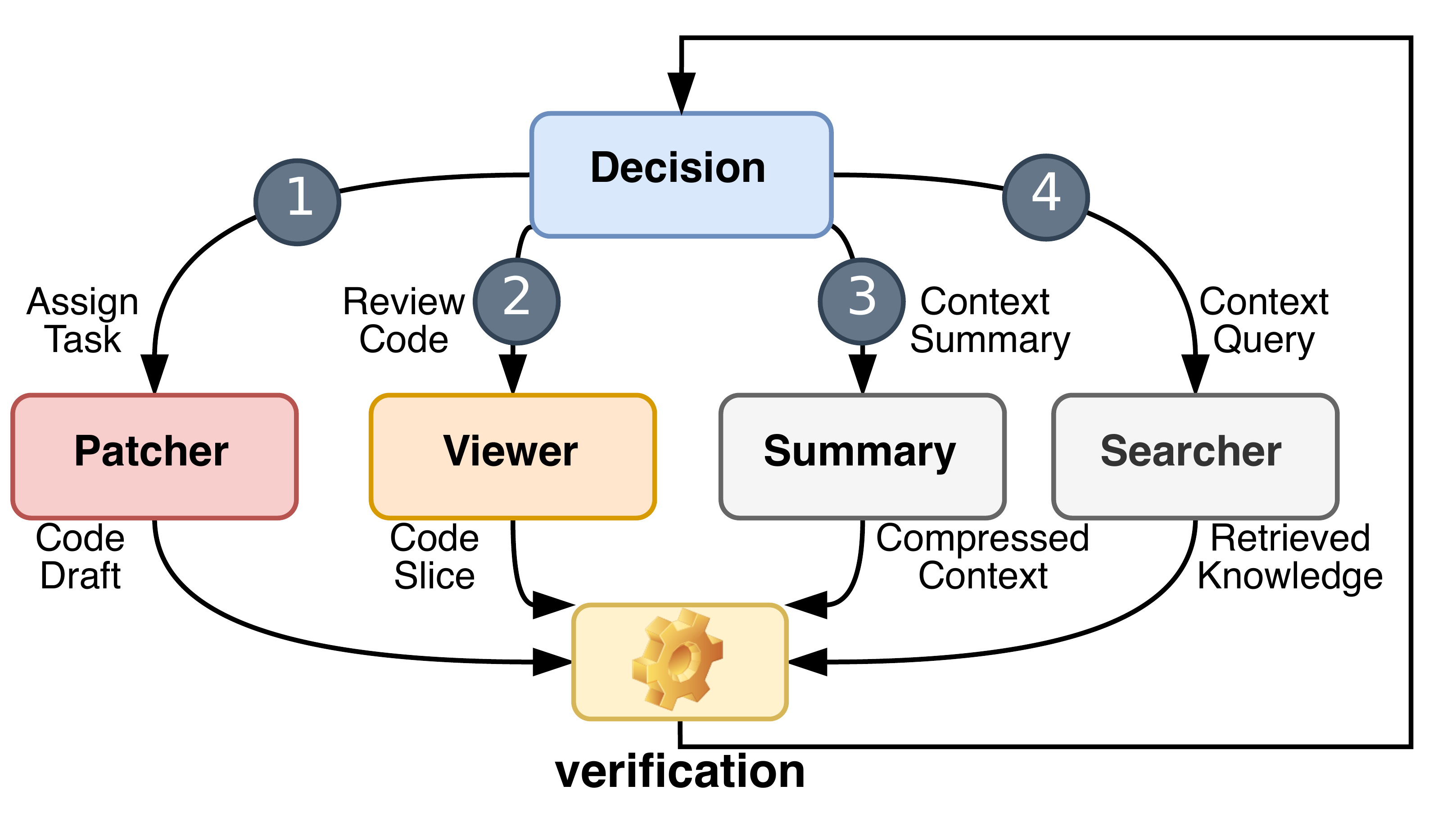}
  \caption{Simplified R$^3$A\cite{r3a} multi-agent workflow example with five agents: \textsc{Decision}, \textsc{Patcher}, \textsc{Viewer}, \textsc{Summary}, and \textsc{Searcher}.}
  \label{fig:agent_case}
\end{figure}
\begin{figure}[t]
  \centering
  \includegraphics[width=0.9\columnwidth]{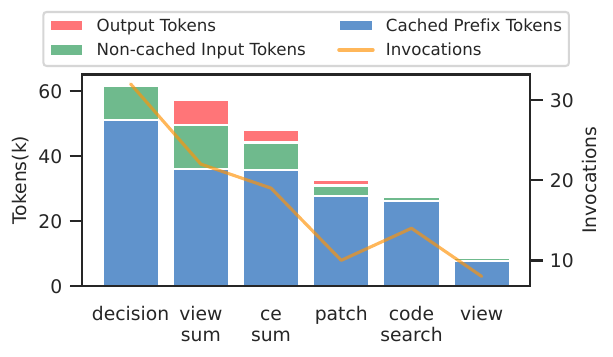}
  \caption{Profiling results of the R3A~\cite{r3a} multi-agent system based on 105 captured calls. Bars: Token usage across agent roles, including non-cached
  input tokens, output tokens, and cached prefix tokens. Line: Invocation  counts 
  across agent roles.}
  \label{fig:mas}
\end{figure}
\begin{figure*}[t]
  \centering
  \includegraphics[width=\textwidth]{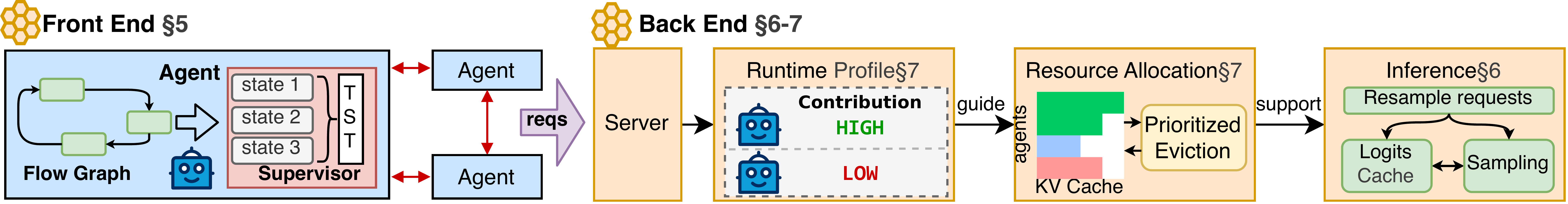}
  \caption{Overview of Hive Multi-Agent Infrastructure}
  \label{fig:overview}
\end{figure*}

\subsection{Cross-Path Redundancy}


At the \textbf{algorithm level}, we observed that test-time scaling, especially hybrid TTS methods, introduces substantial computational redundancy. To illustrate this issue, we use Tree-of-Thoughts~\cite{tree-of-thought} as a representative example. As shown in \autoref{fig:tot_example}(a), ToT organizes reasoning as a branching search process: starting from the input, the LLM expands multiple intermediate states, evaluates them, and continues exploring selected branches until reaching the final output. Here, green states indicate accepted branches retained for continued exploration, whereas red states indicate branches that are pruned during search.


In \autoref{fig:tot_example}(b), we zoom in on the detailed expansion process of \emph{State 4}. Starting from the full context of \emph{State 4}, one branch is first decoded into \emph{State 7} and \emph{State 8}. Obviously, \emph{State 7} and \emph{State 8} share the entire prefix context, and such redundancy can be largely eliminated by existing techniques such as prefix caching\cite{sglang, cached-attn}, which reuse the shared prefix computation across branches.

However, prefix sharing alone is insufficient to fully capture the redundancy in hybrid TTS, since substantial repeated computation still arises beyond the shared context during subsequent decoding. Different branches resampled from the same state often generate overlapping output tokens during decoding, since under similar logits distributions, high-probability tokens remain more likely to be sampled across branches. Nevertheless, each branch still requires a full forward computation at every step to produce these logits again. This repeated recomputation reveals a deeper form of redundancy that is not addressed by prefix reuse alone. We refer to this phenomenon as \textbf{Cross-Path Redundancy}. 

Our solution is introducing the \LogitsCache, which stores the logits of previous inference during decoding. When the same prefix requests a re-sample, the first few decoding steps can be skipped until the sampler takes a different choice on the same logits distribution.

\subsection{Heterogeneous Runtime Characteristics}
At the \textbf{task level}, multi-agent systems improve end-to-end problem solving by decomposing complex tasks into multiple interacting agents, each responsible for a different subproblem or role. However, this execution model also introduces substantially higher request concurrency, more complex inter-agent dependencies, and more heterogeneous resource demands than conventional single-request inference. These characteristics place much stronger pressure on KV cache management, resource allocation, and overall serving throughput.

Unfortunately, current inference\cite{sglang, vllm} engines remain largely agent-agnostic: they schedule and allocate resources at the granularity of individual requests, without awareness of the higher-level agent structure behind them. In multi-agent systems, different agents are assigned distinct roles and therefore naturally handle different types of workloads\cite{SurveyMAS}. To formally characterize this diversity, we performed a detailed profiling on a representative multi-agent system\cite{r3a} with role-specialized agents, including \emph{decision}, \emph{code review}, \emph{error summarization}, and other task-specific agents. For each agent, we measured runtime metrics including invocation frequency, input/output token volume, and dynamic KV cache footprints.

Multi-agent workloads exhibit substantial \textbf{heterogeneous runtime characteristics} across agents, both in invocation frequency and in their impact on end-to-end execution. We choose R$^3$A\cite{r3a}, an open-sourced multi-agent system for RTL program repair, as an example. \autoref{fig:agent_case} shows its simplified workflow. We observed that agents are not invoked uniformly: core agents such as \textsc{Decision}, \textsc{Patcher}, and \textsc{Viewer} are invoked much more frequently, while \textsc{Summary} and \textsc{Search} behave as relatively cold agents. This imbalance implies that KV states generated by different agents do not have equal runtime value. Our profiling results in \autoref{fig:mas} reveal significant heterogeneity in resource consumption across different agent roles. As shown in \autoref{fig:mas}, rather than a uniform distribution, the system's runtime demand is highly skewed across agent roles, with more than 70\% of both total token consumption and invocation frequency concentrated in a subset of agents.

Such heterogeneity becomes especially problematic under memory pressure. In this workflow, the main execution path repeatedly traverses \cOne~and \cTwo, making these agents the dominant contributors to runtime progress. When \textsc{Search} is subsequently invoked to retrieve auxiliary information, the resulting memory spike may push the KV cache toward its capacity limits. The system then needs to invoke \textsc{Summary} to consolidate the retrieved results, requiring additional KV space. Under a traditional policy, such as LRU, eviction is determined solely by generic access order. As a result, KV cache from frequently reused agents such as \textsc{Programmer} or \textsc{Reviewer} may be evicted simply because they are less recent than the newly inserted \textsc{Search} states, even though they are far more likely to be reused in subsequent execution. Such agent-agnostic eviction can significantly degrade overall efficiency in multi-agent workloads.

Based on these observations, \hive~adopts the~\AAS~policy instead of a request-uniform one. In particular, by estimating the contribution of each agent, the system can dynamically adjust its GPU residency budget for KV cache, prioritizing the retention of high-reuse KV states to reduce offloading overhead and improve overall serving efficiency.


\section{Overview}
To address these challenges, we introduce \hive, an inference framework for efficiently supporting both algorithm-level and task-level scaling workloads. \autoref{fig:overview} illustrates the architecture of \hive. \hive~features a descriptive frontend to define the multi-agents(\autoref{sect:frontend}). An agent is described as a coroutine generator, which is internally composed of a flow graph to define the basic behavior and a supervisor to enhance it with test-time scaling algorithms. Then, \hive~proposes a backend to support algorithm- and task-level scaling. First, the server receives per-agent inference requests. Then, the Runtime Profiling determines the contributions of each agent. This information guides Resource Allocation for task-level scaling, together forming the agent-aware scheduling technique(\autoref{sect:schedule}), which prioritizes cache eviction by agent contribution. Afterwards, during the inference process, \hive~introduces the Logits Cache mechanism (\autoref{sect:logits}) to handle resample requests during test-time scaling, which reduces cross-path redundancy at the algorithm level.

\section{Front End}
\label{sect:frontend}

\begin{figure}[ht]
    \centering
    \includegraphics[width=\columnwidth]{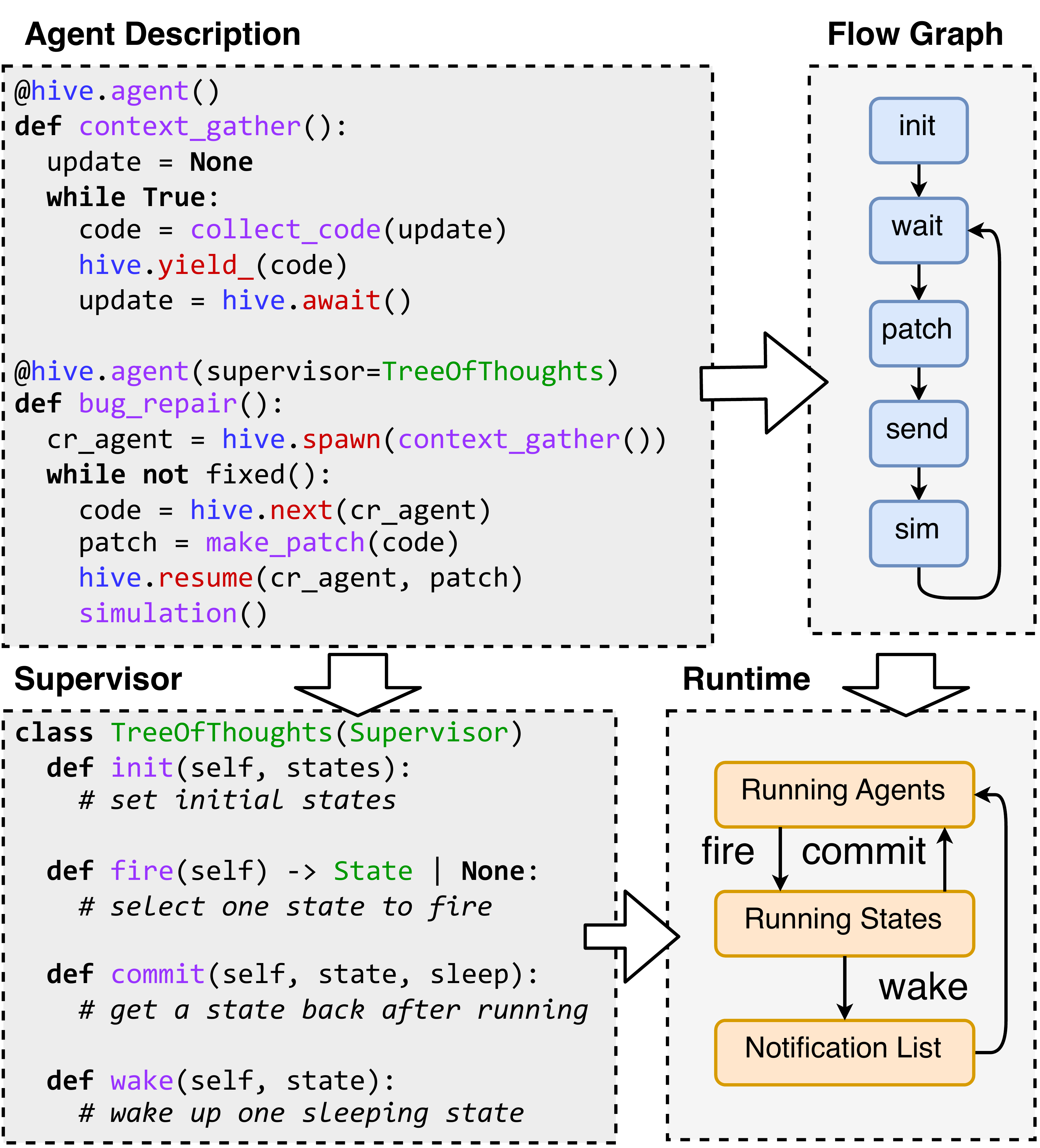}
    \caption{Hive Front End: the Python-based pseudo-code to describe agents and supervisors, the generated flow graph, and the runtime system.}
    \label{fig:frontend}
\end{figure}

\subsection{Agent description and Flow Graph}

\hive~describes the multi-agent system using an abstraction similar to coroutines in Python, to capture their asynchronous nature and simplify programming efforts. The workload of an agent can be described as a coroutine, which is a continuation of control flow while being suspendable to interact with other agents. An example of a program repair agent is shown in \autoref{fig:frontend}. An agent is described as a function with the decorator \textbf{agent}.
The \textit{bug\_repair} agent cooperates with the \textit{context\_gather} agent to help gather the related code snippets and fix the bug. The \textit{bug\_repair} agent first \textbf{spawns} the \textit{context\_gather} agent. Then, in each loop, it receives the gathered code from the \textit{context\_gather} agent by calling \textbf{next}, makes a patch by observing the code, and synchronizes the patch to the \textit{context\_gather} agent by calling \textbf{resume}. The loop continues until the patch passes verification methods like simulation. The \textit{context\_gather} agent is simpler. It keeps gathering code, \textbf{yields} the code to the \textit{bug\_repair} agent, and \textbf{awaits} the update.

Instead of directly executing the functions, \hive~compiles them into flow graphs. As shown in \autoref{fig:frontend}, the flow graph of the \textit{bug\_repair} agent is a finite state machine, exactly describing the behavior of the coroutine. In the graph, a \textbf{state} represents the continuation, transits to another state during one execution step, while inheriting the data and dialogue states for LLMs. This process is similar to what Python interpreters do to a Python async function. 

\subsubsection{Supervisor and Asynchronous Runtime}

\hive~organizes test-time scaling algorithms in the form of \textbf{supervisors} to adapt to asynchronous execution. Each agent is equipped with one supervisor (as a parameter in the \textbf{agent} decorator). Instead of a single function, a test-time scaling algorithm needs to be implemented as an object with four callbacks. As shown in \autoref{fig:frontend}, these callbacks are tightly integrated with the runtime system.
When an agent is spawned, the \textbf{init} function provides the initial states of the flow graph for the algorithm. Then, the runtime will keep querying the supervisor of a running agent whether to \textbf{fire} a state. If a state is fired, then the runtime will put it into the running states set and launch a real Python coroutine to transit the state in its underlying flow graph. When the transition finishes, the new state will be \textbf{committed} to the supervisor. Certain mechanisms are implemented so the runtime can determine whether the transitioned state is sleeping, e.g., waiting for the result from another agent. If the transitioned state would yield data or finish the whole agent, the runtime will look up a notification list to inform the supervisors that some states should \textbf{wake} up, and can be fired if necessary. Such an asynchronous runtime allows running multiple agents in parallel, while allowing agent behavior to be orthogonal to test-time scaling algorithms, enhancing code reuse.

\section{Logits Cache}
\label{sect:logits}

\begin{algorithm}[t]
\caption{Resampling Using Logits Cache}
\label{alg:logits_cache}
\KwIn{Input request $r$, Sampling configuration $\phi$, Replay policy $\pi$}

$s \gets \textsc{GetState}(r)$\;
$\mathcal{Z}_{new} \gets \emptyset$\;

\If{\textsc{HasCachedLogits}$(s)$}{
    $\mathcal{Z}_{cache} \gets \textsc{GetCachedLogits}(s)$\;
    $\mathcal{Y}_{cache} \gets \textsc{GetCachedTokens}(s)$\;

    \If{$\pi = \textsc{StepwiseSampling}$}{
        \ForEach{$z$ in $\mathcal{Z}_{cache}$}{
            $y \gets \textsc{Sample}(z, \phi)$\;
            $r \gets \textsc{AppendToken}(r, y)$\;
            \If{$y$ diverges from cached continuation}{
                \textbf{break}\;
            }
        }
    }
    \ElseIf{$\pi = \textsc{HotspotSampling}$}{
        $\mathcal{H} \gets \textsc{IdentifyHotspots}(\mathcal{Z}_{cache})$\;
        \ForEach{position $i$ in $\mathcal{Z}_{cache}$}{
            \If{$i \in \mathcal{H}$}{
                $y \gets \textsc{Sample}(\mathcal{Z}_{cache}[i], \phi)$\;
            }
            \Else{
                $y \gets \mathcal{Y}_{cache}[i]$\;
            }
            $r \gets \textsc{AppendToken}(r, y)$\;
            \If{$y$ diverges from cached continuation}{
                \textbf{break}\;
            }
        }
    }
}

\If{\textsc{NotFinished}$(r)$}{
    $z \gets \textsc{Prefill}(r)$\;
}
\Else{
    \Return{$r$}\;
}

\While{\textsc{NotFinished}$(r)$}{
    append $z$ to $\mathcal{Z}_{new}$\;
    $y \gets \textsc{Sample}(z, \phi)$\;
    $r \gets \textsc{AppendToken}(r, y)$\;
    \If{\textsc{NotFinished}$(r)$}{
        $z \gets \textsc{Decode}(r)$\;
    }
}

\textsc{UpdateCachedLogits}$(s, \mathcal{Z}_{new})$\;
\Return{$r$}\;
\end{algorithm}
\begin{figure*}[ht]
    \centering
    \begin{subfigure}[c]{0.39\textwidth}
        \includegraphics[width=\textwidth]{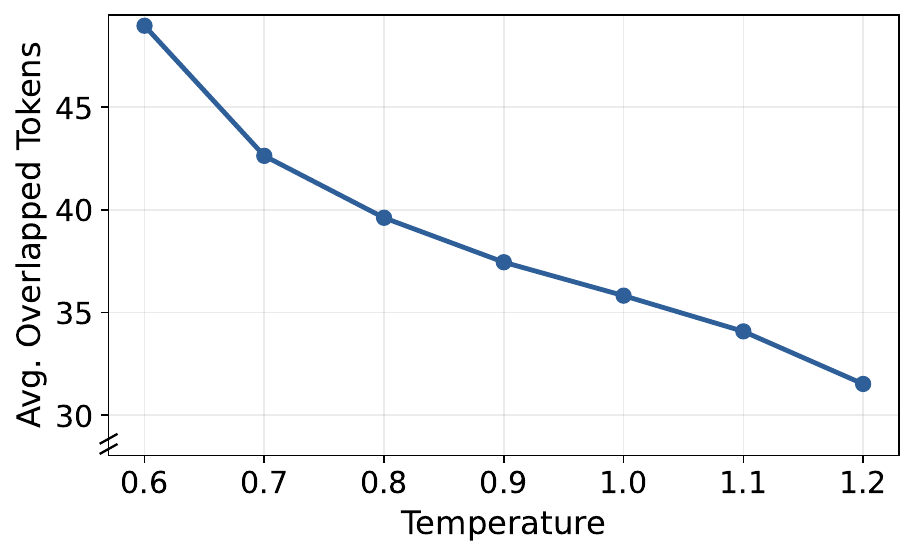}
        \centering
        \caption{Average token overlap with step-wise sampling under different temperatures. }
        \label{fig:tot_overlap}
    \end{subfigure}
    \hfill
    \begin{subfigure}[c]{0.6\textwidth}
        \centering
        \includegraphics[width=\textwidth]{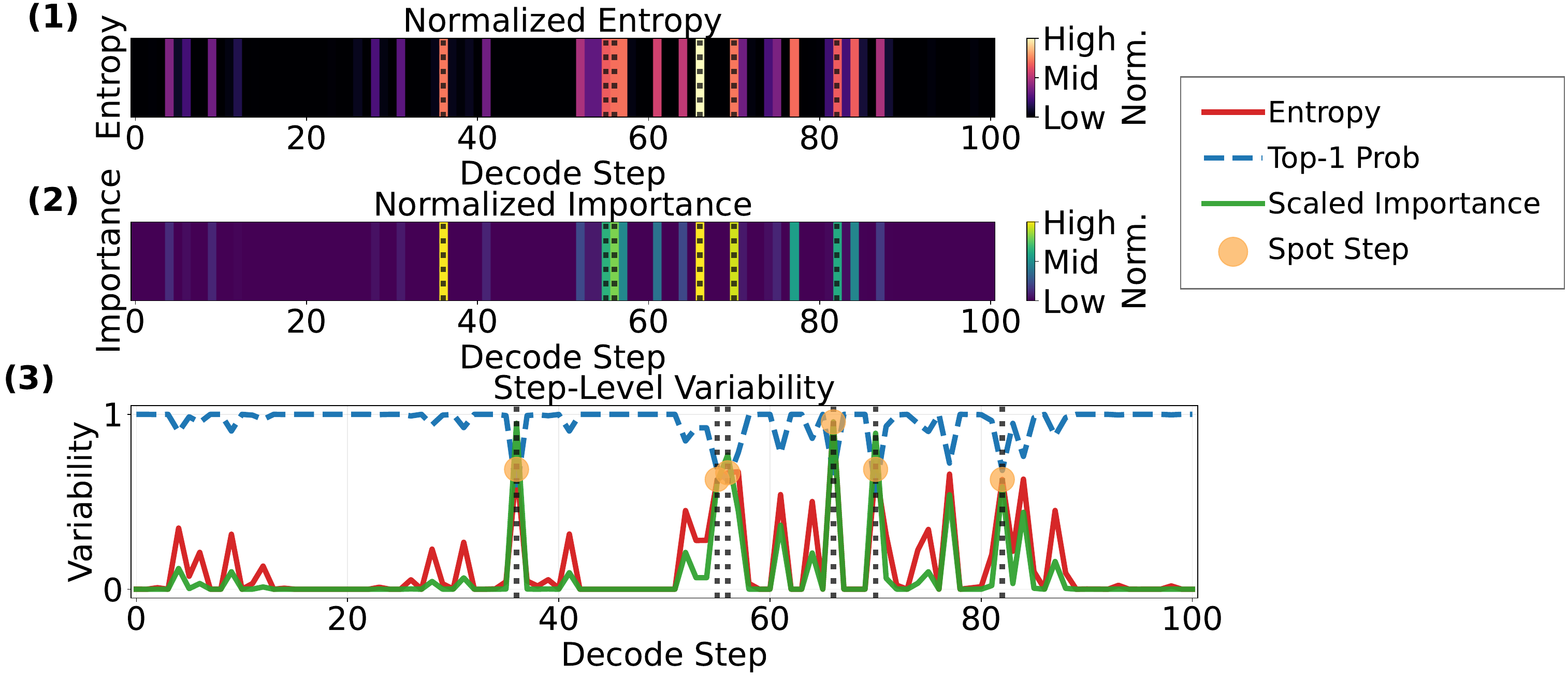}
        \caption{Step-level logit uncertainty on a 100-token decoding trace generated at temperature $0.6$. }
        \label{fig:logits_heat_map}
    \end{subfigure}
    \caption{Profiles for logit cache on R3A~\cite{r3a}. For token overlap, we fix \emph{top-p}=1 and leave \emph{top-k} unconstrained, thereby excluding additional sampling truncation effects from the analysis. For logit uncertainty: (1) Normalized entropy of the token distribution at each decode step. (2) Normalized importance score used by hotspot selection. (3) Step-level variability statistics, including entropy, top-1 probability, scaled importance, and the selected hotspot steps.}
    \label{fig:logitcacheprofile}
\end{figure*}

To reduce the cross-path redundancy in hybrid test-time scaling, we introduce \LogitsCache, a new caching paradigm that reuses intermediate logits across redundant sampling paths. In TTS, the algorithm may repeatedly resample from the same intermediate state. Although these repeated expansions are treated as independent executions, they often traverse highly similar decode trajectories and produce overlapping output tokens. This observation suggests that the next-step logits at a revisited state are not only expensive to regenerate, but also sufficiently informative to be reused for subsequent resampling. \LogitsCache~exploits this property by making such intermediate logits reusable across repeated expansions from the same state.

Formally, at decode step $t$, the transformer-based model\cite{attention} maps the current decoding state to a next-token logits vector with \autoref{equ:transformer}, 
where $x_t$ denotes the current input token at step $t$, $\mathrm{KV}_{<t}$ denotes the cached key-value states of the prefix context, $h_t$ is the hidden representation produced at the current step, and $z_t$ is the logits vector over the vocabulary. Although KV cache avoids recomputing the shared prefix needed to reach the current state, producing $z_t$ still requires a full decode computation at that step, and this cost remains substantial. Taking DeepSeek-V3.2\cite{deepseekv32} as an example, the model comprises 62 layers, so each decode step still incurs a full pass through the model stack to produce the next-token logits. 

During TTS resampling, the requests are executed independently, yet often yield overlapping output tokens due to repeated sampling of high-probability continuations. \hive~removes this redundancy by caching logits from finished requests and reusing them when the same intermediate state is revisited. Specifically, \hive~supports two resampling policies: \emph{Step-wise Sampling} and \emph{Hotspot Sampling}. We describe these two policies next, followed by the overall \LogitsCache~workflow in Algorithm~\ref{alg:logits_cache}.

\begin{equation}
    h_t = f_\theta(x_t, \mathrm{KV}_{<t}),
\qquad
z_t = W_o h_t,
    \label{equ:transformer}
\end{equation}

\subsection{Step-wise Sampling}

Upon a \LogitsCache~hit, \hive~traverses the cached output logits sequentially and samples along the cached trajectory under the sampling configuration. Each sampled token is appended to the request, after which replay advances to the next logits position. This process continues until the cached logits sequence is exhausted or the sampled token diverges from the cached continuation.

This procedure is illustrated in \autoref{fig:tot_example}(c), where the replayed branch directly loads the cached logits from a previously explored state and then performs token-by-token resampling along the cached trajectory. Once the replay diverges from the cached continuation, the tokens sampled so far are concatenated with the original input prefix, after which the system resumes normal prefill and decode.

Step-wise Sampling preserves the original autoregressive execution order and introduces no approximation beyond logits reuse. It therefore retains the full information carried by the cached logits sequence and preserves the original resampling semantics, making it a simple and robust policy for replaying previously explored expansions. In practice, strict token-by-token replay often recovers only a limited reusable output prefix, as shown in \autoref{fig:tot_overlap}, which in turn constrains the achievable reduction in branch-expansion latency.

\subsection{Hotspot Sampling}
To overcome the limited reusable prefix recovered by Step-wise Sampling, \hive~further supports \emph{Hotspot Sampling}. Rather than traversing the cached logits sequence strictly token by token, Hotspot Sampling selectively resamples only a subset of positions that are most likely to alter the downstream decode trajectory.

This design is motivated by an empirical observation: under different temperatures, uncertainty is typically concentrated in only a small fraction of decode steps, rather than being uniformly distributed across the entire sequence. As shown in \autoref{fig:logits_heat_map}(1), only a limited number of positions exhibit pronounced entropy peaks, while most steps remain relatively stable. This suggests that only a small subset of decode steps is responsible for most downstream divergence.

To identify hotspot positions, \hive~computes an importance score for each decode step $t$ from its cached logits $\mathbf{z}_t$. Let $\mathbf{p}_t=\mathrm{softmax}(\mathbf{z}_t)$ be the token distribution at step $t$. We first compute its entropy as in \autoref{equ:entropy} and its maximum token probability as in \autoref{equ:maxprob}, which corresponds to the model's top-1 confidence at step $t$. Then we define the hotspot score as in \autoref{equ:hotspot}, where the $\lambda$ is a time-decay factor. This score favors positions that are both uncertain and weakly peaked, while slightly biasing selection toward earlier positions, whose perturbation is more likely to affect a longer downstream suffix. \hive~selects the top-$k$ steps with the highest scores as hotspot positions and performs explicit resampling only at those steps.

\autoref{fig:logits_heat_map}(2) shows the normalized importance score derived from this formulation, and \autoref{fig:logits_heat_map}(3) overlays the selected hotspot steps with step-level variability statistics. The selected positions align well with regions of high entropy, low top-1 probability, and high importance, indicating that the score effectively captures the subset of steps most likely to affect downstream decoding.

Based on this, \hive~restricts explicit resampling to these hotspot positions, while directly preserving the original sampled tokens at stable positions. This selective policy avoids unnecessary token-by-token replay over regions that are already effectively determined by the logits distribution. As a result, Hotspot Sampling recovers a longer reusable output prefix and further reduces the amount of decode computation required after a cache hit.

\begin{equation}
    H_t = - \sum_{v} p_t(v)\log p_t(v)
    \label{equ:entropy}
\end{equation}
\begin{equation}
    p_t^{\max} = \max_v p_t(v)
    \label{equ:maxprob}
\end{equation}
\begin{equation}
    s_t = H_t (1 - p_t^{\max}) \frac{1}{1+\lambda t}
    \label{equ:hotspot}
\end{equation}

\subsection{Algorithm}
\autoref{alg:logits_cache} describes how \hive~serves a request with \LogitsCache. The algorithm takes as input an incoming request $r$, the sampling configuration $\phi$, and a replay policy $\pi$. Upon receiving $r$, \hive~first derives its current intermediate state $s$ (line 1) and initializes an empty buffer $\mathcal{Z}_{new}$ to record the logits newly generated in this execution (line 2). The engine then checks whether cached logits exist for state $s$ (line 3). If so, it loads both the cached logits sequence and the corresponding cached continuation (lines 4--5).

\hive~supports two replay policies on a cache hit. Under \emph{Step-wise Sampling}, the engine traverses the cached logits sequence in order, samples one token from each logits position, appends it to the request, and stops once the sampled token diverges from the cached continuation (lines 6--11). Under \emph{Hotspot Sampling}, \hive~first identifies a set of hotspot positions from the cached logits sequence (line 13). It then resamples only these hotspot positions, while directly preserving the cached token at non-hotspot positions (lines 14--21). Replay again terminates once divergence from the cached continuation is observed. In this way, Hotspot Sampling avoids unnecessary token-by-token replay over stable regions while still allowing the request to explore alternative continuations at uncertain positions.

After cache replay completes, if the current expansion is still unfinished, \hive~first performs a prefill pass over the replayed tokens to materialize their KV states and obtain the next-step logits (lines 22--23), followed by normal decoding until the current expansion finishes (lines 26--31). Once the expansion completes, \hive~writes the logits sequence generated in this execution back to \LogitsCache~(line 32). Overall, Algorithm~\ref{alg:logits_cache} transforms repeated expansions from the same intermediate state into cache lookup followed by replay and resampling to reduce redundant forward computation.

The \LogitsCache is stored in the main memory instead of the GPU. In practice, to maximize execution efficiency, \hive~executes prefetch sampling asynchronously (lines 3--21). Once \LogitsCache~is updated for the current request (line 32), the runtime immediately triggers a background prefetch task for the same state, overlapping cache update with replay preparation for future revisits. In this way, the latency spent waiting on subsequent sampling is further reduced, thereby shortening the blocking time. 

\section{Agent-Aware Scheduling}
\label{sect:schedule}

\begin{figure*}[t]
  \centering
  \includegraphics[width=\textwidth]{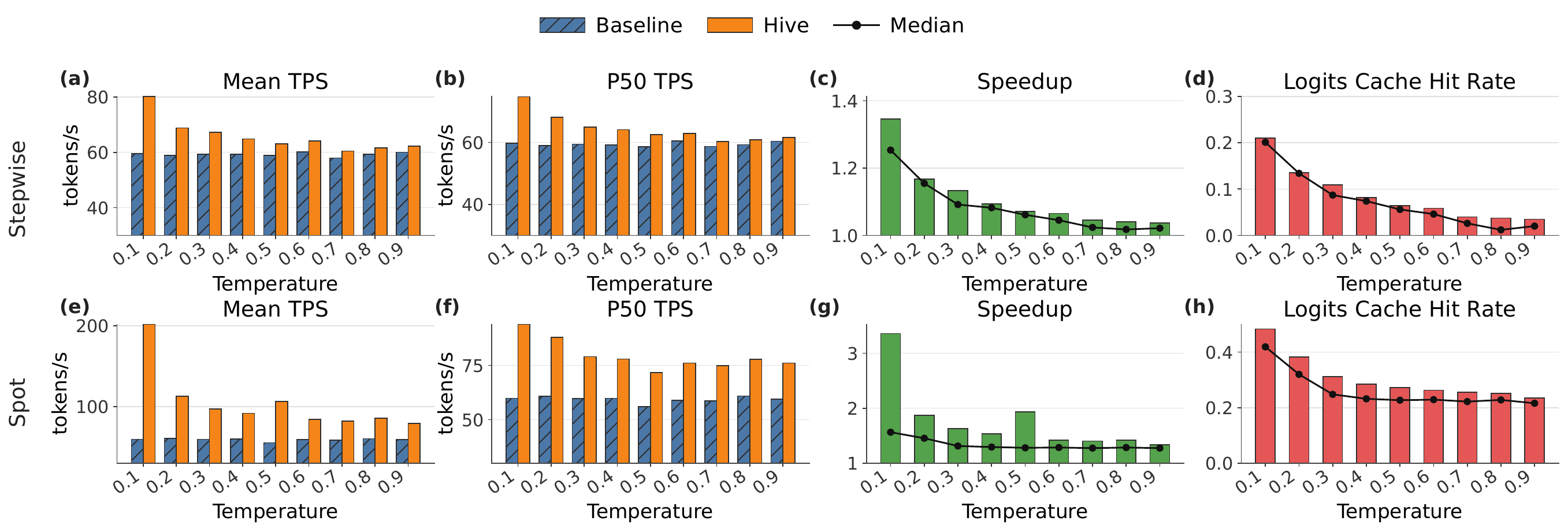}
  \caption{Performance of \LogitsCache~under different temperatures for two replay policies. (a)--(d) show the results for \emph{step-wise sampling}, and (e)--(h) show the results for \emph{hotspot sampling}. The reported metrics include mean TPS, P50 TPS, speedup, and logits-cache hit rate.}
  \label{fig:logits_eval}
\end{figure*}

To address agent-agnostic resource management in multi-agent workloads, we design \AAS, a run-time scheduling mechanism that allocates KV cache residency according to the contribution of each agent. Unlike traditional inference engines that treat requests as independent and uniform entities, \hive~recognizes that requests in multi-agent systems are generated by agents with distinct functional roles and heterogeneous runtime behaviors. Therefore, the scheduler elevates resource management to the agent level, enabling cache residency to be guided by the actual utility each agent brings to the workflow. In the following, we first introduce our Shapley-based contribution modeling to quantify the importance of each agent. We then detail how the scheduler leverages these contribution scores to perform dynamic KV cache eviction, improving the overall efficiency of the multi-agent execution.

\subsection{Agent Contribution Modeling}
A key limitation of existing inference systems is their agent-agnostic resource allocation. In complex multi-agent workflows, requests are generated by agents with highly heterogeneous roles, execution patterns, and cache-sharing potential. Consequently, a naive request-uniform policy fails to account for the disproportionate impact certain agents have on the overall system efficiency and task progress. To bridge this gap, \hive~models each agent's marginal contribution by drawing on the contribution calculation principle of the Shapley value from game theory~\cite{shapley_value}.

Specifically, for a set of active agents $\mathcal{A}$, Hive assigns each agent $a \in \mathcal{A}$ a holistic contribution score. This score integrates two dimensions of agent behavior: 
\begin{itemize}[leftmargin=*]
  \item \textbf{Intrinsic Utility ($\widehat{U}_{a}$):} Quantifies the standalone importance of an agent's individual task progress. 
  \item \textbf{Collaborative Utility ($\widehat{C}_{a}$):} Captures the systemic benefit derived from shared-cache reuse that the agent facilitates for other members in the coalition. 
\end{itemize}
The final contribution score for agent $a$ is defined as in \autoref{equ:contribution},
where $\widehat{U}_{a}$ and $\widehat{C}_{a}$ denote the normalized intrinsic and collaborative scores. In our implementation, we empirically set $\alpha = 0.4$, prioritizing collaborative reuse to maximize the throughput gains from shared KV cache states across the agentic workflow.

\noindent\textbf{Intrinsic Utility.}
The intrinsic utility of an agent captures its standalone runtime importance. Intuitively, an agent should receive a higher score if it is invoked more frequently, achieves better cache efficiency, exhibits stronger concurrency. \hive~models this utility as the product of four normalized factors as in \autoref{equ:intrinsic}.
In this formulation, the activity term ($A_a$) and workload term ($W_a$) together capture the \emph{inference momentum} of an agent, with $A_a$ scaling with invocation frequency and $W_a$ measuring the aggregate input and output token volume. The efficiency term ($E_a$) captures \emph{resource productivity} through KV cache reuse, favoring agents with stronger temporal locality. The concurrency term ($Q_a$) summarizes parallel demand using both average and peak concurrency, thereby prioritizing agents that place greater sustained pressure on GPU execution resources.

\noindent\textbf{Collaborative Utility.}
Beyond intrinsic utility, an agent's marginal contribution also depends on the additional benefit it induces for the rest of the workflow. In \hive, a key source of such marginal gain is shared-cache reuse across agents. Agents that create more reusable cache state for others should therefore receive higher collaborative utility scores.

To capture this effect, \hive~measures collaborative utility from the bidirectional shared-cache interaction graph induced at runtime. For each pair of agents $(a,b)$, \hive~tracks the amount of shared-cache reuse between them and converts it into an edge weight defined as in \autoref{equ:weight},
where $\mathrm{S}_{a,b}$ denotes the total shared-token count and $\mathrm{Event}_{a,b}$ represents the reuse events between the pair. The square-root term prevents large token volumes from dominating the score, while the logarithmic factor rewards repeated collaboration.

The aggregate collaborative utility of agent $a$ is then defined as in \autoref{equ:collaborate}.
By summing both the reusable state contributed to others and the reusable state received from others, this formulation provides a computationally efficient approximation of the collaborative component of the Shapley value

\begin{equation}
    \text{Score}(a) = \alpha \cdot \widehat{U}_{a} + (1 - \alpha) \cdot \widehat{C}_{a}
    \label{equ:contribution}
\end{equation}
\begin{equation}
    U_a = A_a \cdot W_a \cdot E_a \cdot Q_a
    \label{equ:intrinsic}
\end{equation}
\begin{equation}
    C_a = \sum_{b\neq a} w(a,b) + \sum_{b\neq a} w(b,a)
    \label{equ:collaborate}
\end{equation}
\begin{equation}
    w(a,b) = \sqrt{\mathrm{S}_{a,b}} \cdot \log(1+\mathrm{Event}_{a,b})
    \label{equ:weight}
\end{equation}

\subsection{Agent-Aware KV Cache Scheduling}

To address the limitation of heterogeneous runtime characteristics, \hive~leverages the agent contribution scores to guide KV cache scheduling at runtime. Rather than relying on generic eviction policies, \hive~prioritizes KV cache according to the contribution of the agent that owns them. KV cache associated with higher-contribution agents is retained on GPU whenever possible, since evicting them is more likely to hurt overall execution efficiency. Under memory pressure, eviction is therefore biased toward KV cache belonging to lower-contribution agents. The contribution score is calculated during run-time profiling, which is updated in each round of requests. The front-end registers the existence of each agent and sends their IDs along with the requests, so the backend can track down each agent. A sliding window is applied to prevent a breaking change in agents' behaviors.
In this way, \hive~aligns KV cache residency with the structural importance of agents in the workflow, improving memory utilization and preserving the states that matter most for efficient multi-agent execution. 

\section{Evaluation}

\begin{figure}[t]
  \centering
  \includegraphics[width=\columnwidth]{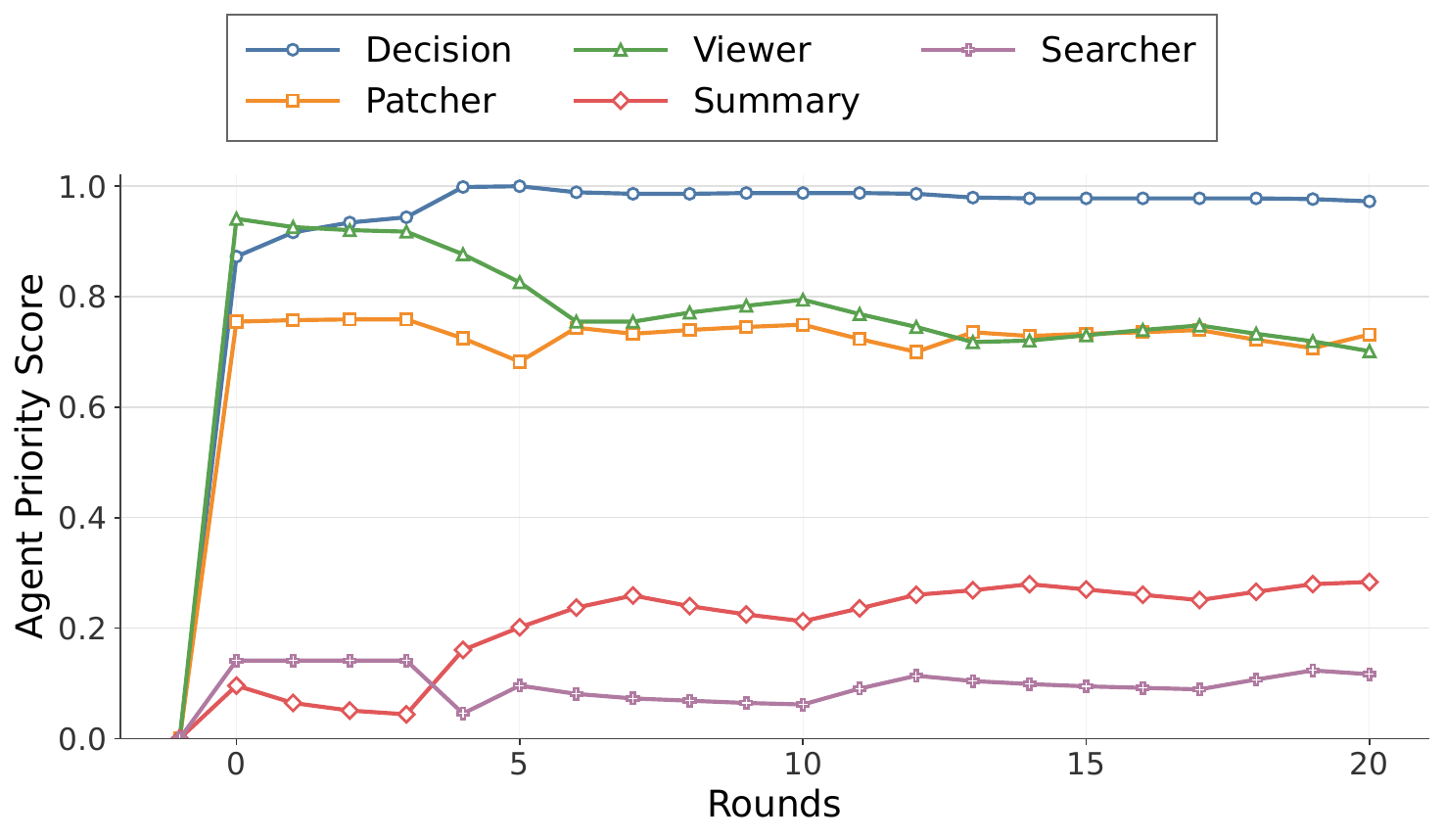}
  \caption{Trajectory of agent priority scores over 21 rounds. For each round, scores are normalized across all agents.}
  \label{fig:agent_score_trace}
\end{figure}

We evaluate \hive~from two perspectives, corresponding to the two scaling dimensions that it targets. For \emph{algorithm-level scaling}, we evaluate \LogitsCache~on Tree-of-Thoughts workloads to measure how effectively it reduces replay overhead during repeated resampling. For \emph{task-level scaling}, we evaluate the \emph{Multi-Agent Scheduler} on multi-agent workloads to study whether agent-aware resource allocation improves execution efficiency over existing scheduling policies. Together, these experiments demonstrate how \hive~provides unified runtime support for both replay-efficient reasoning and scalable multi-agent execution.

\subsection{Experimental Setup}
\noindent\textbf{Platform.}
The experiments are conducted on a server with 8$\times$NVIDIA A100 GPUs. It is equipped with an Intel Xeon Gold 6348 CPU @ 2.60 GHz. The software stack includes CUDA Toolkit 11.5, PyTorch 2.9.1, Python 3.12.13, and SGLang 0.5.6. We implement \hive~on top of SGLang and evaluate it under the same runtime environment throughout. All experiments are run on a single GPU.

\noindent\textbf{Model.}
We use Qwen3-8B\cite{qwen3} as the base model throughout the evaluation. We choose this model because it provides a practical open-weight setting for controlled replay, resampling, and multi-agent serving experiments. While large-scale experiments are absent due to practical reasons, our technique is theoretically expected to become more effective at larger scales, as the cost saved by a cache hit grows with the model scale.

\subsection{Logits Cache Evaluation}
\noindent\textbf{Workload.}
We evaluate \LogitsCache~on repeated resampling workloads induced by hybrid test-time scaling using the Tree-of-Thoughts dataset\cite{tree-of-thought}. Specifically, a request first completes normal decoding and records its logits sequence, after which later requests that revisit the same intermediate state perform replay-based resampling from cached logits, which directly matches the target scenario of \LogitsCache. The workload is constructed from the Hugging Face 24 Game dataset\cite{24games}, which contains 1.36k instances of four-number arithmetic puzzles with annotated solutions. From this dataset, we sample 100 instances as the evaluation workload, whose average prompt length is 64.52 tokens.

\noindent\textbf{Baseline.}
Our baseline disables \LogitsCache~and recomputes the full logits sequence for every replayed request. To ensure a fair comparison, the baseline and \hive~use the same KV cache configuration and retain identical prefix-state reuse behavior; the only difference is whether the replayed request can reuse cached logits or must regenerate them through normal decoding.

\noindent\textbf{Decoding Configuration.}
Unless otherwise specified, the baseline and \hive~use identical decoding configuration, including temperature, top-$p$, top-$k$, and \texttt{max\_tokens}, for a fair comparison. We consider two replay settings. Since sampling is inherently stochastic, repeated runs may produce outputs of different lengths, making the results harder to reproduce and compare fairly. To reduce this variation, we fix the output length for every request in both settings. Following \cite{dualpath}, which reports an average output length of 429 tokens for agentic workloads, we set the fixed output length to 500 tokens. For \emph{hotspot sampling}, decoding positions with an importance score greater than 0.6 are treated as \emph{spot nodes}. We vary the temperature to study how the benefit of \LogitsCache~changes across different replay conditions.

\noindent\textbf{Metrics.}
For all reported metrics, we exclude requests with a 100\% Logits Cache hit ratio, since such requests would distort the measurement results. In total, we filter out 30 requests. We then report the following metrics for \LogitsCache~during resampling requests: \textbf{MEAN TPS}, the average tokens per second; \textbf{P50 TPS}, the median number of tokens per second; \textbf{Speedup}, the average per-request speedup; \textbf{Logits-cache hit ratio}, the fraction of decoding positions in a replayed request whose logits are served from the cache instead of being recomputed by the model.

\noindent\textbf{Performance.}
We first evaluate the replay TPS reduction achieved by \LogitsCache~under different replay policies and sampling temperatures. \autoref{fig:logits_eval} presents the performance of \LogitsCache~under step-wise and hotspot replay across a range of temperatures. Overall, \LogitsCache~consistently improves replay TPS compared with the baseline, with the largest gains appearing at low temperatures. The benefit of \LogitsCache~is expected to grow with prompt length, since a longer prefill context substantially increases the computational cost of each subsequent decoding step, whereas \LogitsCache~can bypass these expensive forward computations by directly sampling from cached logits distributions. This advantage is particularly salient in agentic workloads, where context windows are often much longer than those in conventional settings. For instance, DualPath~\cite{dualpath} reports an average input length of 32.7K tokens for agentic requests, a regime in which logits-based reuse is expected to deliver even greater benefits.

Under \emph{step-wise sampling}, both mean and P50 TPS are substantially improved when the temperature is low. As temperature increases, replayed branches tend to diverge earlier from the cached continuation, which reduces the amount of reusable logits and weakens the gain.\LogitsCache~provides consistent TPS improvement across all evaluated temperatures. Under \emph{hotspot sampling}, the improvement is even more pronounced, yielding much larger speedups in the low-temperature regime. Compared with strict step-wise replay, hotspot replay operates exclusively on high-importance positions, which allows it to validate and accept larger token segments once a cache reuse is successful. 

The token-acceptance behavior of hotspot replay also explains the statistical pattern observed in \autoref{fig:logits_eval}(g). Since the reuse of a single hotspot node can trigger the acceptance of a cascading sequence of subsequent tokens, the average speedup under this policy exhibits higher variance. As a result, while the mean speedup in \autoref{fig:logits_eval}(g) shows noticeable fluctuations, the median remains relatively stable and aligns with the expected theoretical trends. 

In conclusion, across all evaluated temperatures, \LogitsCache~achieves an average speedup of $1.11\times$ under step-wise replay and $1.76\times$ under hotspot replay, with corresponding average Logits Cache hit rates of $8.6\%$ and $30.4\%$, respectively.

\begin{table}[t]
    \centering
    \caption{Comparison of KV-cache hit rates and token eviction between the baseline scheduler and the agent-aware scheduler. Hotspot and non-hotspot hit rates measure the KV-cache hit rate over hotspot and non-hotspot tokens, respectively, while Evicted tokens denotes the total number of tokens whose KV states are evicted during execution. We set page size = 1 by default.}
    \label{tab:online_eval_rounds}
    \small
    \setlength{\tabcolsep}{3.5pt}
    \begin{tabular}{clccc}
      \toprule
      Round & Method & Hotspot & Non-hotspot & Evicted \\
            &        & hit rate & hit rate    & tokens \\
      \midrule
      7  & \textsc{Baseline} & 0.903 & 0.311 & 26557 \\
      7  & \textsc{HIVE}     & 0.935 & 0.310 & 18549 \\
      \midrule
      21 & \textsc{Baseline} & 0.935 & 0.414 & 81147 \\
      21 & \textsc{HIVE}     & 0.967 & 0.311 & 65598 \\
      \bottomrule
    \end{tabular}
\end{table}
\subsection{Multi-Agent Scheduler Evaluation}
\noindent\textbf{Workload.}
To evaluate the scheduler under practical conditions, we construct a benchmark multi-agent workflow based on the execution pattern shown in \autoref{fig:agent_case}. The workload integrates five specialized agents: \textsc{Decision}, \textsc{Patcher}, \textsc{Viewer}, \textsc{Summary}, and \textsc{Searcher}. Specifically, \textsc{Decision} acts as the central coordinator that manages workflow transitions. \textsc{Patcher} and \textsc{Viewer} serve as the hotspot agents in the system. In contrast, \textsc{Summary} is responsible for context consolidation and state compression, whereas \textsc{Searcher} performs external retrieval and can introduce transient yet substantial memory pressure. Together, these heterogeneous agent roles create highly uneven runtime demands, providing a rigorous testbed for evaluating \hive's ability to balance competing agent utilities and optimize KV cache residency. To ensure reproducibility, we fix the agent invocation relationships as a static workflow during evaluation, so that all compared scheduling policies are tested under the same execution structure.

\noindent\textbf{Memory Setting.}
We set \texttt{mem-fraction-static} to 0.23 to emulate a memory-constrained serving environment, thereby amplifying cache contention and making the impact of different eviction policies more pronounced. This value is chosen based on the model size and the prompt length used in the benchmark.

\noindent\textbf{Baseline.}
We compare \hive's agent-aware policy against the original LRU-based cache eviction policy in SGLang.

\noindent\textbf{Performance.}
As shown in \autoref{tab:online_eval_rounds}, the agent-aware scheduler consistently improves KV cache efficiency over the baseline. At round 7, \hive~increases the hotspot hit rate from 0.903 to 0.935, while achieving a 1.43$\times$ reduction in evicted tokens. By round 21, \hive~further improves the hotspot hit rate from 0.935 to 0.967, while achieving a 1.24$\times$ reduction in evicted tokens. These results show that the agent-aware policy is able to better preserve KV states associated with critical execution paths under memory pressure. Notably, this improvement mainly comes from prioritizing hotspot tokens, rather than uniformly increasing the hit rate for all tokens, which is consistent with the design goal of favoring high-contribution agents during cache eviction.

The runtime score traces of individual agents further confirm this behavior, shown in \autoref{fig:agent_score_trace}. We observe that agent scores evolve consistently with their runtime invocation patterns over time, rather than remaining static or uniform across agents. This indicates that our contribution formulation can effectively model the relative contribution of different agents during execution, enabling the scheduler to capture agent heterogeneity online and adjust scheduling priorities accordingly.

Overall, across different rounds, \hive~reduces the hotspot miss rate by $33\%$--$51\%$ over the baseline, while reducing evicted KV tokens by $19.2\%$--$30.2\%$, demonstrating the effectiveness of agent-aware scheduling under memory pressure.

\section{Related Work}

\subsection{Task-Level Scaling}
KVFlow~\cite{kvflow} and TokenCake~\cite{tokencake} study KV-cache management for \emph{multi-agent} LLM serving. KVFlow proposes a workflow-aware mechanism that constructs an \emph{Agent Step Graph} offline from the workflow structure, and assigns each agent a graph-derived timestamp to guide KV eviction and prefetching, thereby improving prefix reuse in structured agentic workloads. TokenCake~\cite{tokencake} presents a KV-cache-centric serving framework for multi-agent applications, providing a DSL to describe inter-agent dependencies as a DAG and using this structure to drive a \emph{Time Scheduler} and a \emph{Space Scheduler} for execution coordination and KV-cache allocation across agents. ThunderAgent~\cite{thunderagent} mainly focuses on \emph{single-agent} execution. It introduces a program-aware scheduling framework that jointly manages KV cache, program state, and tool resources. In contrast, \hive~focuses on runtime contribution-aware scheduling for dynamic multi-agent serving, allocating KV-cache resources according to the observed runtime behavior and relative contribution of different agents.

\subsection{Algorithm-Level Scaling}
Recent work has started to improve algorithm-level scaling from a systems perspective. For example, FastTTS~\cite{FastTTS} studies execution scheduling and memory management for verifier-guided test-time scaling, aiming to improve serving efficiency under more complex inference workloads. In contrast, \hive~ focuses on a different source of inefficiency in hybrid TTS settings: redundant re-sampling over shared probability states across reasoning paths. \hive~addresses this issue through logits-level reuse, which is complementary to prior system support for TTS.

\section{Conclusion}

Current LLM infrastructures mainly scale at the model level and system level. We propose \hive~, a multi-agent infrastructure for algorithm- and task-level scaling. Experiments show that our proposed \LogitsCache~achieves an average speedup of $1.11\times$--$1.76\times$ for re-sampling, and \AAS~reduces the hotspot miss rate by $33\%$--$51\%$.


\bibliographystyle{ACM-Reference-Format}
\bibliography{references}

\end{document}